\documentclass{llncs}
\usepackage[utf8]{inputenc}
\usepackage{amsmath}
\usepackage{amssymb}
\usepackage{graphicx}
\usepackage{float}
\usepackage[caption=false]{subfig}
\usepackage{algorithm}
\usepackage[noend]{algorithmic}
\usepackage[hidelinks]{hyperref}

\newcommand{\Z}{\mathbb{Z}}

\newcommand{\R}{\mathbb{R}}
\newcommand{\landau}{\mathcal{O}}

\begin{document}
\title{Streaming Algorithm for Euler Characteristic Curves of Multidimensional Images}
\author{Teresa Heiss\inst{1}\fnmsep\inst{2} \and Hubert Wagner\inst{1}}
\institute{IST Austria, Am Campus 1, 3400 Klosterneuburg, Austria\\
	\email{teresa.heiss@ist.ac.at}, \email{hubert.wagner@ist.ac.at} \and Vienna University of Technology, Karlsplatz 13, 1040 Vienna, Austria}
\maketitle
\vspace*{-0.19cm}
\begin{abstract}
We present an efficient algorithm to compute Euler characteristic curves of gray scale images of arbitrary dimension. In various applications the Euler characteristic curve is used as a descriptor of an image.

Our algorithm is the first streaming algorithm for Euler characteristic curves. The usage of streaming removes the necessity to store the entire image in RAM. 
Experiments show that our implementation handles terabyte scale images on commodity hardware. Due to lock-free parallelism, it scales well with the number of processor cores. Our software---CHUNKYEuler---is available as open source on Bitbucket.

Additionally, we put the concept of the Euler characteristic curve in the wider context of computational topology. In particular, we explain the connection with persistence diagrams.
\end{abstract}
\section{Introduction}
\label{sec:introduction}
The Euler characteristic curve is a powerful tool in image processing~\cite{Gonzalez_Wintz_1977}. It has been used in a variety of fields including astrophysics\footnote{The astrophysics community refers to the Euler characteristic curve as the genus.}~\cite{gott,rhoads1994genus}, medical image analysis~\cite{segonne2007geometrically,odgaard1993quantification}, and image processing in general~\cite{Richardson201499,Snidaro20031533}.
Its wide applicability stems from simplicity and efficient computability.

However, with the advances in image acquisition technology, there is need to handle very large images. For example the state-of-the-art micro-CT scanner Skyscan 1272 creates images of size $14450\times14450\times2600$ with 14-bit precision. Therefore, a single scan yields more than half a trillion voxels. It is also possible to combine multiple scans of the same object which further multiplies the size of data. As loading the resulting multi-terabyte image into RAM of a commodity computer is infeasible, a streaming approach is needed. 

We present the first streaming algorithm for computing Euler characteristic curves.\footnote{We review related work at the end of the paper.} Our algorithm divides a multidimensional image into chunks that fit into RAM, calculates the Euler characteristic curves for each chunk separately and merges them in the end. Since these chunks can be made arbitrarily small, commodity hardware can be used to compute Euler characteristic curves of arbitrarily large images. The fact that the chunks are not dependent on each other makes lock-free parallelism possible. 

For defining the Euler characteristic curve we first need to explain what the Euler characteristic is. There are two ways to define the Euler characteristic and the Euler-Poincaré formula states that they are both equivalent. For discrete two-dimensional surfaces, like a triangulation of a sphere or a torus, the definitions are: first, the number of vertices minus the number of edges plus the number of faces; second, the number of connected components minus the number of tunnels plus the number of voids. Originally the Euler characteristic was defined for the surface of a convex polyhedron where it always equals two. To see this, consider that such a surface consists of one connected component, no tunnels and one void.

The equivalence between these two definitions seems to be the reason for the usefulness of the Euler characteristic: It captures global topological structures---like holes---although it can be computed locally---by adding up vertices, edges and faces. 

\begin{figure}[!h]
		\centering
		\subfloat[Example image]{
			\centering
			\includegraphics[width=0.147\linewidth]{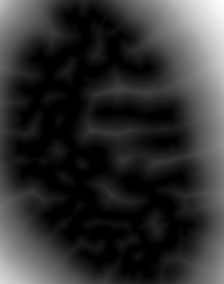}
			\label{fig:exampleImageSubfigure}
		}\ 
		\subfloat[$t=0$\newline$\chi(T_0)=8$]{
			\centering
			\includegraphics[width=0.147\linewidth]{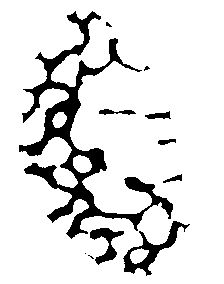}
			\label{fig:t0}
		}
		\subfloat[$t=12$\newline$\chi(T_{12})=-17$]{
			\centering
			\includegraphics[width=0.147\linewidth]{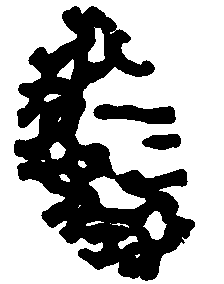} 
			\label{fig:t1}
		}
		\subfloat[$t=25$\newline$\chi(T_{25})=-7$]{
			\centering
			\includegraphics[width=0.147\linewidth]{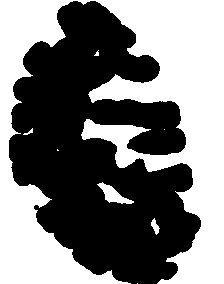} 
			\label{fig:t2}
		}
		\subfloat[$t=38$\newline$\chi(T_{38})=-3$]{
			\centering
			\includegraphics[width=0.147\linewidth]{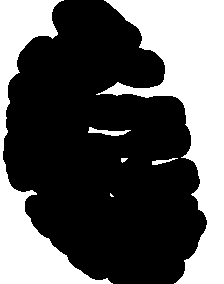} 
			\label{fig:t3}
		}
		\subfloat[$t=52$\newline$\chi(T_{52})=1$]{
			\centering
			\includegraphics[width=0.147\linewidth]{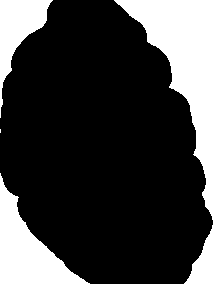} 
			\label{fig:t4}
		}\\
		\subfloat[Euler characteristic curve of the example image]{
			\centering
			\includegraphics[trim=1.9cm 0cm 3.14cm 0.28cm,clip,
			width=\textwidth,height=74.94pt]{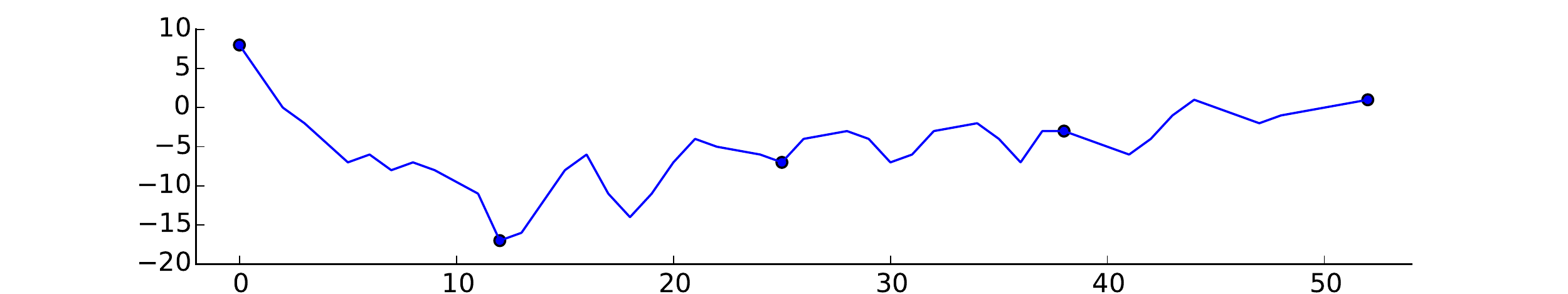}
			\label{fig:eulercurve}
		}
		\caption[Definition of the Euler characteristic curve]{Definition of the Euler characteristic curve illustrated with an image showing a 2D slice of the distance transform of a segmented bone. Subfigures~\subref{fig:t0}--\subref{fig:t4} show thresholded images $T_t$ for different thresholds $t$. For $t=38$, there is 1 component and 4 holes, hence $\chi=-3$. Starting from $t=52$, there are no holes, so $\chi$ stabilizes at 1.}
		\label{fig:explanation_ECC}
\end{figure}

The Euler characteristic curve of an image is the vector of Euler characteristics of consecutive thresholded images. We illustrate this for the example image of a bone\footnote{We thank Reinhold Erben and Stephan Handschuh from Vetmeduni Vienna for providing micro-CT scans of rat vertebrae.} in Fig.~\ref{fig:exampleImageSubfigure} with values ranging from 0 (black) to 255 (white). The Euler characteristic curve of this image maps each $t \in \{0,1,...,255\}$ to the Euler characteristic of the set of pixels with gray value smaller or equal to $t$. Figure~\ref{fig:explanation_ECC} illustrates this process. This concept can be extended to more general settings, e.g., images with floating point gray values (see Sec.~\ref{sec:algorithm}).
\section{Theoretical Background}
\label{sec:background}
\begin{figure}[!h]
	\centering
	\includegraphics[width=0.25\textwidth]{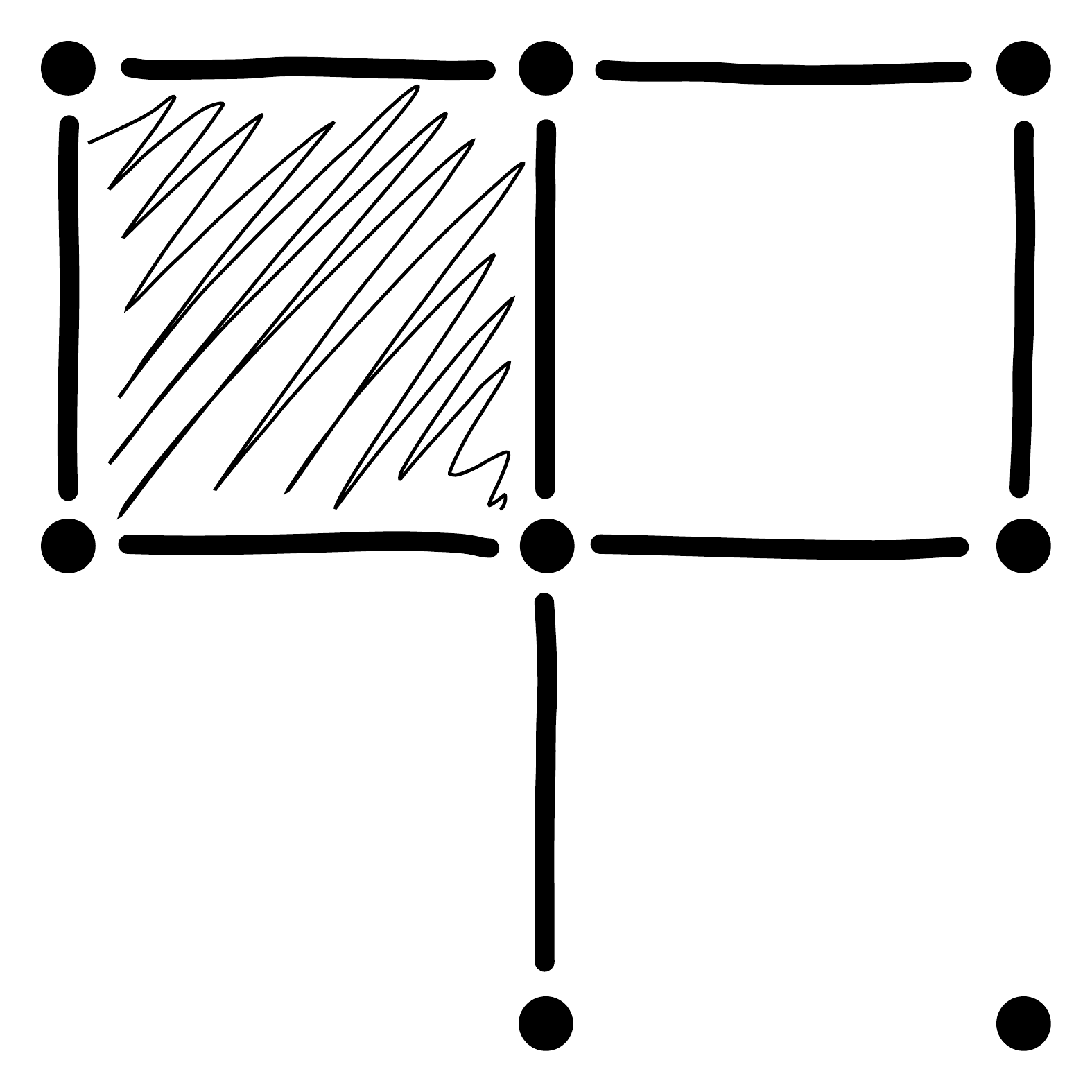}
	\caption{Example of a two-dimensional cubical complex of embedding dimension two with one square, eight edges and eight vertices.}
	\label{fig:example_complex}
\end{figure}

We give basic definitions needed in Sec.~\ref{sec:algorithm} using the language borrowed from computational topology~\cite{kaczynski2004computational}. With this we can provide precise definitions in arbitrary dimension and explain the connection between the Euler characteristic curve and other topological descriptors.
\paragraph{Cubical Cell.}
A $k$-dimensional \textbf{cubical cell} (short: \textbf{cell}) $c$ of embedding dimension $d$ is defined as the Cartesian product of intervals and singletons:
\begin{equation*}
c:=I_1 \times I_2 \times \dots \times I_d
\end{equation*}
where exactly $k$ of the sets $(I_i)_{i\in\{1, 2, \dots,d\}}$ are intervals of the form $I_i=[a_i,a_i+1]$ with integers $a_i\in\Z$ and the remaining $d-k$ sets are singletons $I_i=\{b_i\}$ with integers $b_i\in\Z$. A zero-dimensional cell is called a \textbf{vertex}, a one-dimensional cell an \textbf{edge}, a two-dimensional cell a \textbf{square}, a three-dimensional cell a \textbf{cube}.
\paragraph{Face.}
A cell $c_1$ of embedding dimension $d$ is called a \textbf{face} of a cell $c_2$ of embedding dimension $d$ if $c_1$ is a subset of $c_2$.
\paragraph{Cubical Complex.}
A $p$-dimensional \textbf{cubical complex} (short: \textbf{complex}) of embedding dimension $d$ is a finite set of cubical cells of embedding dimension $d$ such that
\begin{enumerate}
	\item The faces of each cell are also elements of the complex 
	\item The intersection of any two cells is also an element of the complex\footnote{The second condition is implied by the first condition since we allow only consecutive integers as interval endpoints in the definition of cells.}
\end{enumerate}
where $p$ is the highest dimension of all cells in the complex. The complexes that appear in our algorithm always fulfill $p=d$. Figure~\ref{fig:example_complex} shows an example of a cubical complex.
\paragraph{Filtration of Cubical Complexes.}
A sequence of complexes $K_1,K_2,\dots,K_m$ is called a \textbf{filtration} if the complexes are monotonically increasing: $K_1\subseteq K_2\subseteq\dots\subseteq K_m$.
\paragraph{Sublevel Set Filtration.}
Let $K$ be a cubical complex. A cell that is not a face of any other cell than itself is called a \textbf{maximal cell}. Let $f:M\to\R$ be a function, where $M$ is the set of maximal cells. This function $f$ can be extended to a function $\tilde{f}:K\to\R$ defined on all cells: 
\begin{align*}
\tilde{f}\colon K &\to \R\\
c &\mapsto \tilde{f}(c):=
\begin{cases} 
f(c) & \mbox{if } c\in M\\
\displaystyle\min_{\substack{m\in M\\ c\subseteq m}}f(m) & \mbox{otherwise.}
\end{cases}
\end{align*}

For each $t \in \R$ the \textbf{sublevel set} $\tilde{f}^{-1}\left((-\infty,t]\right)$ of this extended function is the set of cells that are a face of at least one maximal cell with $f$-value smaller or equal to $t$. As $K$ consists of only a finite number of cells, $\tilde{f}$ can only have a finite number of different function values $\{t_1, t_2,\dots,t_m\}$.
The sublevel sets $\tilde{f}^{-1}\left((-\infty,t_1]\right),\dots,\tilde{f}^{-1}\left((-\infty,t_m]\right)$ form a filtration of cubical complexes---the \textbf{sublevel set filtration} induced by the function $f$.

To see this, notice that the definition of $\tilde{f}$ implies that for each $t\in\R$ the sublevel set $\tilde{f}^{-1}\left((-\infty,t]\right)$ is a cubical complex: all faces of a cell $c$ belong to the same sublevel set as $c$. Furthermore the sublevel sets are monotonically increasing $\tilde{f}^{-1}\left((-\infty,t_1]\right) \subseteq \tilde{f}^{-1}\left((-\infty,t_2]\right) \subseteq \dots \subseteq \tilde{f}^{-1}\left((-\infty,t_m]\right)$. Therefore, the sublevel sets form a filtration.
\paragraph{Consecutive Thresholded Images as Sublevel Set Filtrations.}
A $d$-dimensional gray scale image with $n_1\times n_2 \times \dots \times n_d$ voxels\footnote{Throughout this paper we use ``voxel'' as multidimensional generalization of ``pixel''.} can be interpreted as a $d$-dimensional cubical complex $K$ of embedding dimension $d$ with a function $f$ on its maximal cells: for each voxel index $(i_1,\dots,i_d)$, $i_1 \in \{1,\dots,n_1\},\dots,i_d\in \{1,\dots,n_d\}$ the corresponding voxel position is represented by the $d$-dimensional cell $c_{i_1,\dots,i_d}$ of embedding dimension $d$:
\begin{equation*}
c_{i_1,\dots,i_d} := \left[i_1 - 1;i_1\right]\times\left[i_2 - 1;i_2\right]\times\dots\times\left[i_d - 1;i_d\right] \enspace .
\end{equation*}
The cubical complex $K$ is defined as the set of all these cells $c_{i_1,\dots,i_d}$ along with all their faces. The function $f$ maps each maximal cell $c_{i_1,\dots,i_d}$ to the gray value of the voxel with index $(i_1,\dots,i_d)$. The sublevel set filtration\footnote{Another interpretation of voxel data is via the dual complex (voxels become vertices) using the lower star filtration. The way we use appears more natural in image processing context. The two approaches yield similar but not necessarily identical Euler characteristic curves.} induced by the function $f$ is formed by consecutive thresholdings of the image.\footnote{Defining cells as products of \emph{closed} intervals implies $(3^d-1)$-connectivity for the voxels of the thresholded images. This corresponds to 8-connectivity for 2D images.}
\paragraph{Euler Characteristic.}
The \textbf{Euler characteristic} $\chi$ of a $p$-dimensional complex $K$ of embedding dimension $d$ is defined as 
\begin{equation*}
\chi(K):=\sum_{k=0}^{p}(-1)^k n_k
\end{equation*}
where $n_k$ is the number of $k$-dimensional cells in $K$.

The Euler-Poincaré formula states 
\begin{equation*}
\chi(K)=\sum_{k=0}^{p}(-1)^k n_k =\sum_{k=0}^{d-1}(-1)^k\beta_k
\end{equation*}
where $\beta_k$ is the $k$th Betti number (the number of $k$-dimensional holes). For a formal definition of cubical homology and the involved Betti numbers, see~\cite{kaczynski2004computational}. In three-dimensional space, $\beta_0$ is the number of connected components, $\beta_1$ is the number of tunnels and $\beta_2$ is the number of voids.
\paragraph{Euler Characteristic Curve.}
The \textbf{Euler characteristic curve} $e$ of a filtration of cubical complexes $K_1\subseteq K_2\subseteq\dots\subseteq K_m$ is the vector 
\begin{equation*}
e=\left(\chi(K_1),\chi(K_2),\dots,\chi(K_m)\right)\enspace .
\end{equation*}
This vector can also be interpreted as a function
\begin{align*}
e\colon \{1,2,\dots,m\} &\to \Z\\
t &\mapsto \chi(K_t) \enspace ,
\end{align*}
which is used to visualize the Euler characteristic curve as in Fig.~\ref{fig:eulercurve}. 
\paragraph{Euler Characteristic Curve of an Image.}
We already saw that for an arbitrary gray scale image the sequence of consecutive thresholded images is a filtration---the sublevel set filtration. The Euler characteristic curve of this filtration is the \textbf{Euler characteristic curve of an image}. 
\paragraph{Connection to Other Topological Descriptors.}
We want to put the above considerations in the wider context of computational topology. Two popular topological descriptors of a filtration are Betti curves and persistence diagrams~\cite{edelsbrunner2010computational}, which both capture information about holes at different thresholds. 

For each hole in the image the persistence diagram tracks the first and last threshold at which the hole occurs. The $k$th Betti curve, which counts the $k$-dimensional holes at each threshold, is easily computable from the persistence diagram. Furthermore, the alternating sum of the Betti curves yields the Euler characteristic curve. Therefore, the Euler characteristic curve summarizes a persistence diagram, but it can be computed locally.

The usefulness of the Euler characteristic curve suggests that the two richer descriptors may also be useful in image processing. However, large images are out of reach of the currently available persistence diagram software. For now, the Euler characteristic curve remains the only feasible option.
\section{Algorithm}
\label{sec:algorithm}
The input for our algorithm is a gray scale image of arbitrary dimension $d$ with $n$ voxels. The output is the Euler characteristic curve of this image, as defined in Sec.~\ref{sec:background}.
\subsubsection{Range of Values.}
In Sec.~\ref{sec:background} the function $f$ maps to $\R$. However, in practice the gray values of an image are in a predefined range, usually $\{0,1,\dots,255\}$ or $\{0,1,\dots,65535\}$. If the range contains negative numbers, it can be shifted so that it starts from zero. For this reason we focus on ranges of the form $\{0,1,\dots,m-1\}$ with a positive integer $m$. A version of our algorithm that can handle floating point values will be discussed at the end of this section.
\subsubsection{Tracking the Changes.}
It is suboptimal to compute the Euler characteristic for each threshold separately, as already noted in~\cite{Snidaro20031533}. To avoid redundant computations we track the changes between consecutive thresholds. More precisely, we determine how each voxel contributes to the change in Euler characteristic. Therefore we first compute a vector of changes in Euler characteristic (VCEC) $\left(a_0, a_1, \dots, a_{m-1}\right)$ whose entries $a_t := \chi \left( \tilde{f}^{-1}\left(\left(-\infty,t\right]\right) \right) - \chi \left( \tilde{f}^{-1}\left(\left(-\infty,t-1\right]\right) \right)$ are the difference between the Euler characteristics of two consecutive thresholded images.\footnote{where $\chi \left( \tilde{f}^{-1}\left(\left(-\infty,-1\right]\right) \right)=\chi(\emptyset)=0$.} The Euler characteristic curve is then:
\begin{align*}
&\left(a_0, a_0+a_1, \dots, \sum_{t=0}^{m-1}a_t\right) =\\ =&\left( \chi \left( \tilde{f}^{-1}\left(\left(-\infty,0\right]\right) \right), \chi \left( \tilde{f}^{-1}\left(\left(-\infty,1\right]\right) \right), \dots, \chi \left( \tilde{f}^{-1}\left(\left(-\infty,m-1\right]\right) \right) \right) \enspace .
\end{align*}

When changing from one thresholded image $\tilde{f}^{-1}\left(\left(-\infty,t-1\right]\right)$ to the next $\tilde{f}^{-1}\left(\left(-\infty,t\right]\right)$, all voxels with gray value $t$ are included, along with all their faces that have not already been included at a previous threshold. We say that these new faces are introduced by these new voxels. More precisely, a face $c$ of a voxel $v$ is \textbf{introduced} by $v$ if all other voxels $w$ that have $c$ as a face fulfill one of the following two conditions:
\begin{enumerate}
	\item $f(w)>f(v)$
	\item $f(w)=f(v)$ and $w\succeq v$\enspace, \label{enum:order}
\end{enumerate}
where $\succeq$ is any total order of the voxel positions, e.g., the lexicographical order\footnote{In lexicographical order a voxel at position $(i_1,\dots,i_d)$ succeeds a voxel at position $(j_1,\dots,j_d)$ if $i_k>j_k$ for the first $k$ where $i_k$ and $j_k$ differ.}. The output of our algorithm is independent of the chosen total order. However, it is necessary to require $w\succeq v$ in condition \ref{enum:order} to ensure that each face is introduced by a unique voxel.

\begin{figure}[!h]
	\centering
	\subfloat[The cubical complex consists not only of the voxels but also of their faces (red).]{
		\centering
		\includegraphics[width=0.468\linewidth]{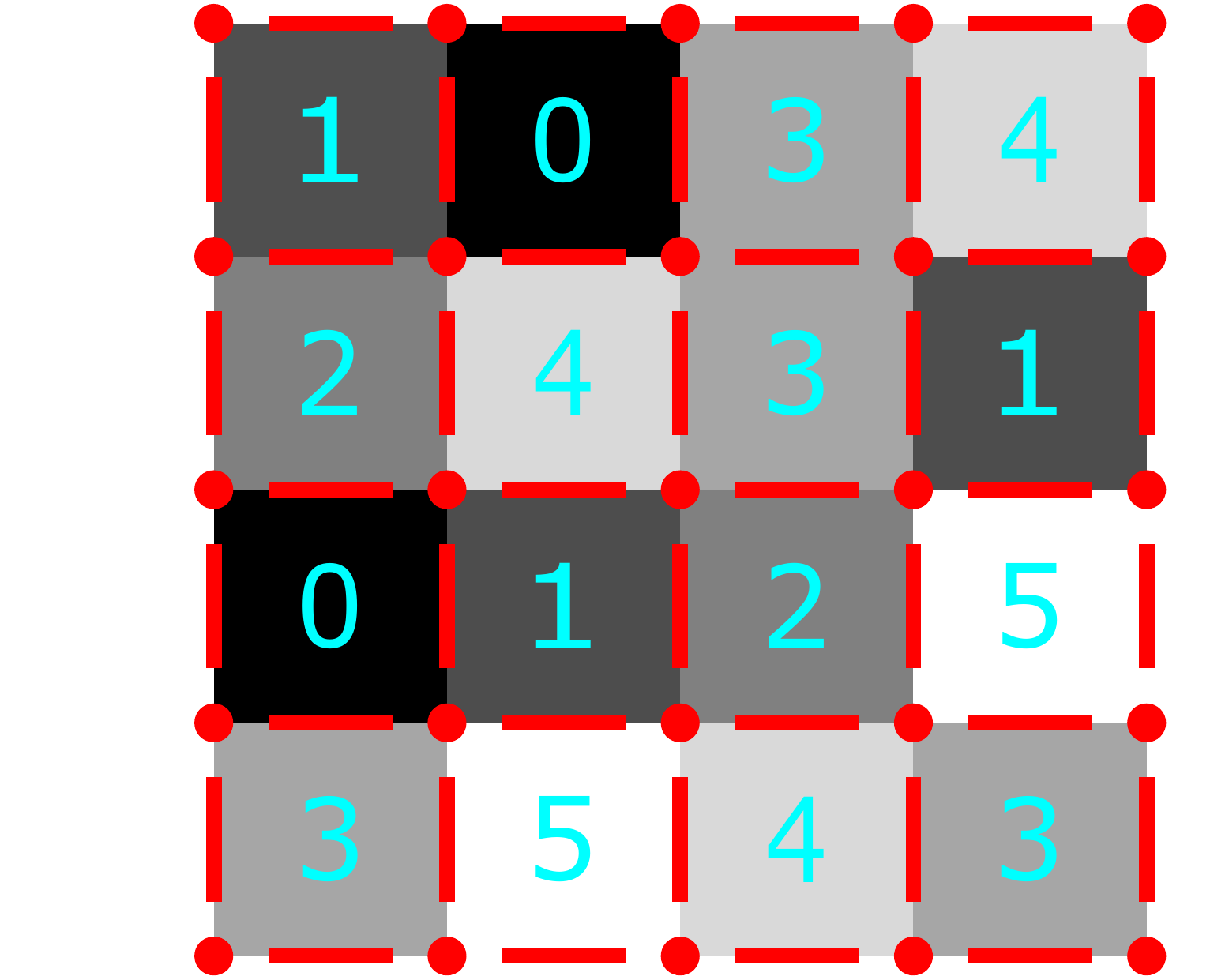}
		\label{fig:before_decomposition}
	}\quad
	\subfloat[Block decomposition shows which voxels introduce which faces, e.g., the upper left voxel introduces 3 edges and 2 vertices.]{
		\centering
		\includegraphics[width=0.468\linewidth]{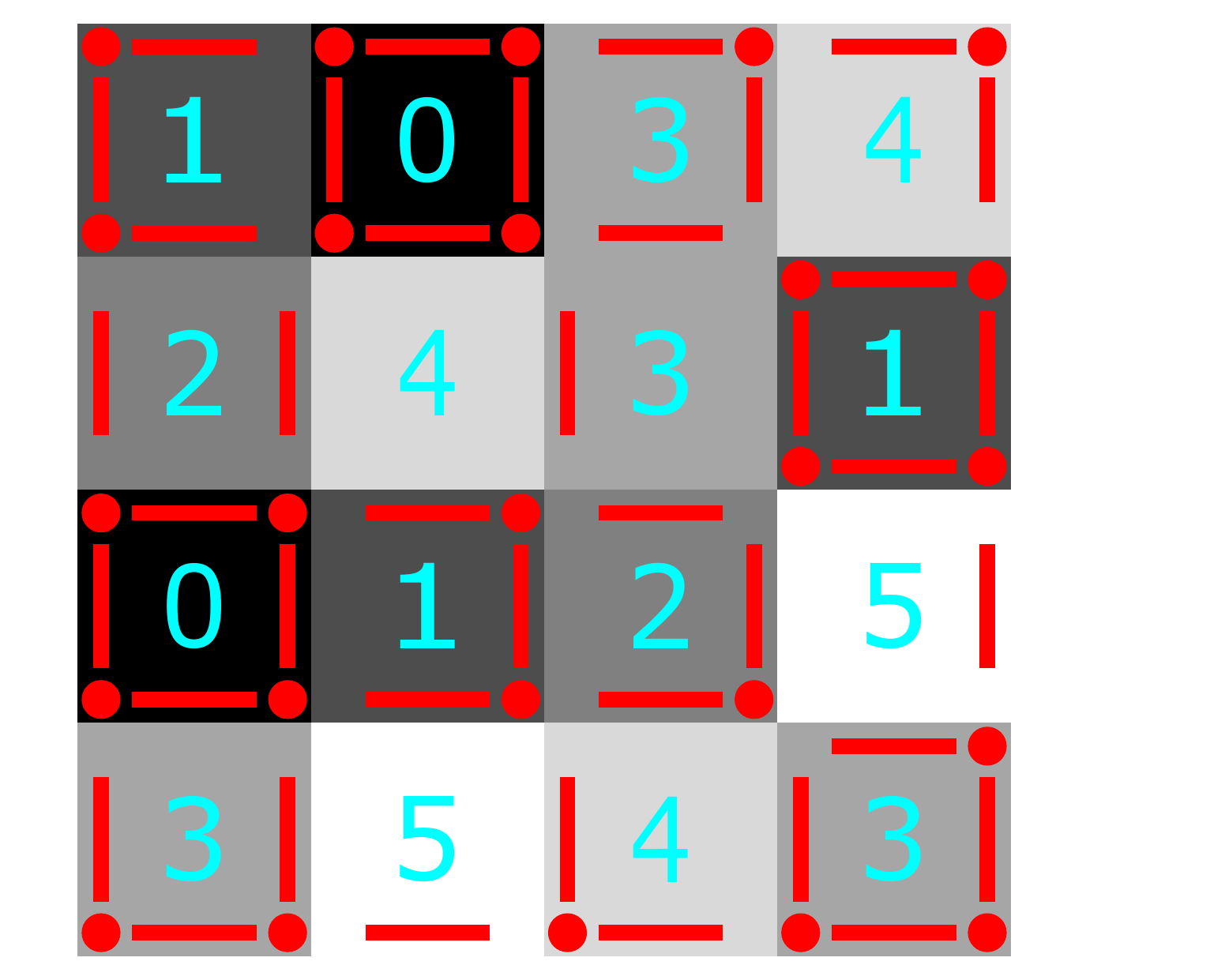} 
		\label{fig:after_decomposition}
	}
	\caption[Decomposition into blocks]{This is an illustration for the block decomposition.}
	\label{fig:blocks}
\end{figure}

Now we can decompose the cubical complex of the input image into blocks such that each block contributes to exactly one change of threshold. A block consists of a voxel together with all the faces it introduces. Which faces are introduced by a certain voxel is determined only by the gray values of the voxel's $3^d-1$ neighbors. We exploit this locality in the design of our parallel streaming algorithm. Figure~\ref{fig:blocks} shows the decomposition of a two-dimensional example image into blocks.
\subsubsection{Storage.}
For the computations we store only the gray values of the voxels. The geometric information and the adjacency relations between cells are implicit in the voxel grid and calculated locally whenever needed. Similarly the function $\tilde{f}$ and the block decomposition it induces are never explicitly stored. Apart from storing the result vector, the memory overhead is essentially zero.
\subsubsection{Streaming.}
\begin{figure}[!h]
	\centering
	\includegraphics[width=0.861\textwidth]{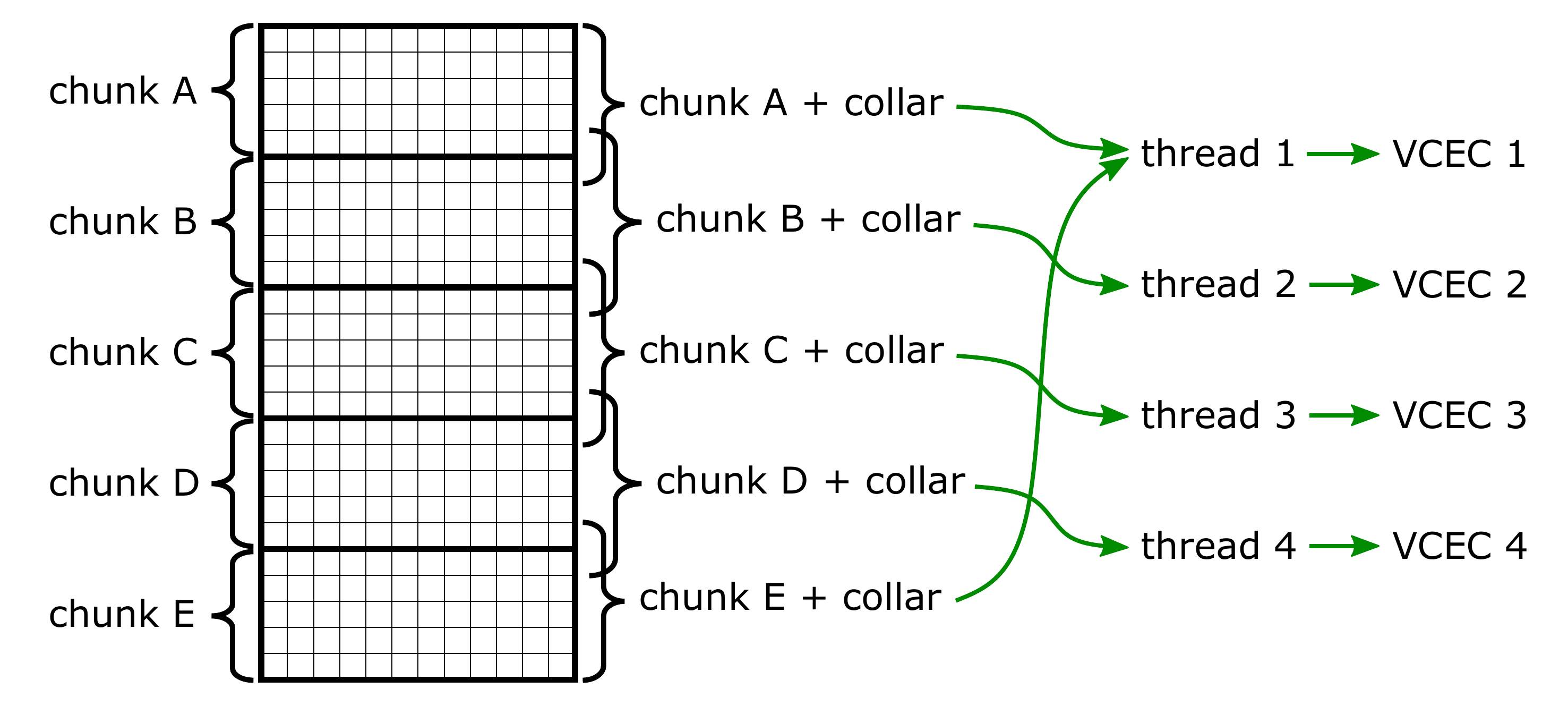}
	\caption{Whenever a worker thread is free it loads one chunk along with a one voxel thick collar into its space in RAM and computes the VCEC of this chunk.}
	\label{fig:cores}
\end{figure}

If the entire image does not fit into RAM, we divide it into chunks that fit into RAM. In our implementation we use a simple strategy: an image of size $n_1 \times n_2 \times \dots \times n_d$ is divided into $c$ chunks of size $\frac{n_1}{c} \times n_2 \times \dots \times n_d$ (see Fig.~\ref{fig:cores}, left). As these correspond to contiguous memory regions, streaming the chunks from a single input file is easy. We then separately compute the VCEC for each chunk either sequentially or in parallel.
\subsubsection{Parallel Computations.}
For parallelism we use a thread pool. Each worker thread is assigned memory for a single chunk and one initially empty VCEC vector. One task is to read a chunk from disk, update the worker thread's VCEC vector by the VCEC of this chunk and discard the chunk. At any given time at most $w$ chunks reside in RAM, where $w$ is the number of worker threads. Because different worker threads work with disjoint memory regions we achieve lock-free parallelism. The underlying data structure for the collection of VCECs is a vector of vectors, called \verb|euler_changes|.
\subsubsection{Processing one Chunk.}
Along with the chunk we read a one voxel thick collar surrounding it. This way we have access to all neighbors of the voxels in the chunk (see Fig.~\ref{fig:cores}). With this information we compute the VCEC of this chunk as specified in Algorithm~\ref{alg}.
\begin{algorithm}
	\caption{Computing the VCEC}
	\label{alg}
	\begin{algorithmic}[1]
		\REQUIRE One chunk of an image along with a one voxel thick collar surrounding it and \verb|current_thread|, the index of the worker thread processing this chunk.
		\ENSURE An updated version of the vector \verb|euler_changes|[\verb|current_thread|] which is the VCEC of all chunks this thread has processed.
		\FORALL{voxels $v$ in the chunk}
		\STATE{$t$ = gray value of $v$}
		\STATE change $=0$
		\FORALL{faces $c$ introduced by $v$}
		\IF{dim($c$) is even}
		\STATE change$++$
		\ELSE
		\STATE change$--$
		\ENDIF
		\ENDFOR
		\STATE{\verb|euler_changes|[\verb|current_thread|][$t$] += change} \label{line}
		\ENDFOR
		\STATE remove chunk and collar from RAM
	\end{algorithmic}
\end{algorithm}
\vspace*{-0.4cm}
\subsubsection{Post-Processing.}
In the end, when all chunks have been processed, a single thread sums up the VCECs yielding the VCEC of the whole image, which is a vector $\left(a_0, a_1, \dots, a_{m-1}\right)$. The Euler characteristic curve is then computed as $\left(a_0, a_0+a_1, \dots, \sum_{t=0}^{m-1}a_t\right)$.
\subsubsection{Analysis.}
To analyze the complexity we remind that $d$ is the dimension of the image, $n$ is the number of voxels, $m$ is the number of gray values\footnote{The input size is $\log_2(m)n$.} and $w$ is the number of worker threads. We introduce a new variable $s$, the number of voxels per chunk including the collar.

Assuming perfect parallelization, the worst case running time of our algorithm is $\landau(\frac{3^dn}{w}+mw)$ because for each voxel we visit all its neighbors and sequentially post-process the VCECs. We analyze the practical scaling behavior in Sec.~\ref{sec:experiments}. As the dimension $d$ is usually small (mostly 2 or 3), the exponential term is usually not a problem in practice.

For each worker thread, we need $s$ integers of $\log_2(m)$ bits to store the gray values of a chunk. Additionally, the \verb|euler_changes| data structure consists of $wm$ 64-bit integers. Therefore, the total storage is $\log_2(m)ws+64wm+\landau(1)$ bits in RAM. By decreasing the chunk size $s$, the dominant part, $\log_2(m)ws$, can be made arbitrarily small. Because of this, our algorithm works for arbitrarily large images on commodity hardware.
\subsubsection{Other Ranges of Gray Values.}
When the range of gray values is not of the form $\{0,1,\dots,m-1\}$---for example for floating point values---one option is to use a hash map to store the \verb|euler_changes|. If the number of different input values approaches $n$, the output size dominates the overall storage and the advantage of a streaming approach disappears. In this situation, it is preferable to transform (i.e., round, scale, shift) the input values to obtain a range of the form $\{0,1,\dots,m-1\}$. Running our standard algorithm on the transformed input yields the same result as transforming the domain of the Euler characteristic curve computed for the original data.
\section{Experiments}
\label{sec:experiments}
We implemented the above algorithmic scheme in C++14. We made experiments on two different machines: a laptop with Intel core i5-5200U CPU with two physical cores clocked at 2.2\,GHz with 8\,GB of RAM and a workstation with Intel Xeon E5645 CPU with 12 physical cores clocked at 2.4\,GHz and 72\,GB of RAM. Table~\ref{tab:constanttime} shows the running time and memory usage for different 3D input images ran on the laptop. We use images from a standard data set\footnote{Most of the images are available at \url{www.byclb.com/TR/Muhendislik/Dataset.aspx}}, see~\cite{wagner2011efficient,delgado2015skeletonization}. The names' suffixes distinguish between 8- and 16-bit precision images. The last column shows that the running time is linear in $n$ and does not depend on the content of the image. 
\begin{table}[!h]
	\centering
	\caption{Running time and memory usage.}
	\label{tab:constanttime}
	\begin{tabular}{llrrrr}
		\hline
		name          & size            &   million voxels &   memory[MB] &   time[s] &   time[s]/million voxels \\
		\hline
		prone16         & 512$\times$512$\times$463    &       121.4 &         70.4 &      24.7 &             0.20 \\
		xmastree16 & 512$\times$499$\times$512    &       130.8 &         72.9 &      26.7 &             0.20 \\
		vertebra16    & 512$\times$512$\times$512    &       134.2 &         74   &      28.4 &             0.21 \\
		random8 & 512$\times$512$\times$512    &       134.2 &         51   &     28.9 &           0.22 \\
		random8 & 1024$\times$1024$\times$1024 &      1073.7 &         93.1 &    236.7 &           0.22  \\
		random8 & 2048$\times$2048$\times$2048 &      8589.9 &        261.6 &   1767.1 &           0.21 \\
		\hline
	\end{tabular}
\end{table}

\begin{figure}[!h]
	\centering
	\includegraphics[width=0.84\textwidth]{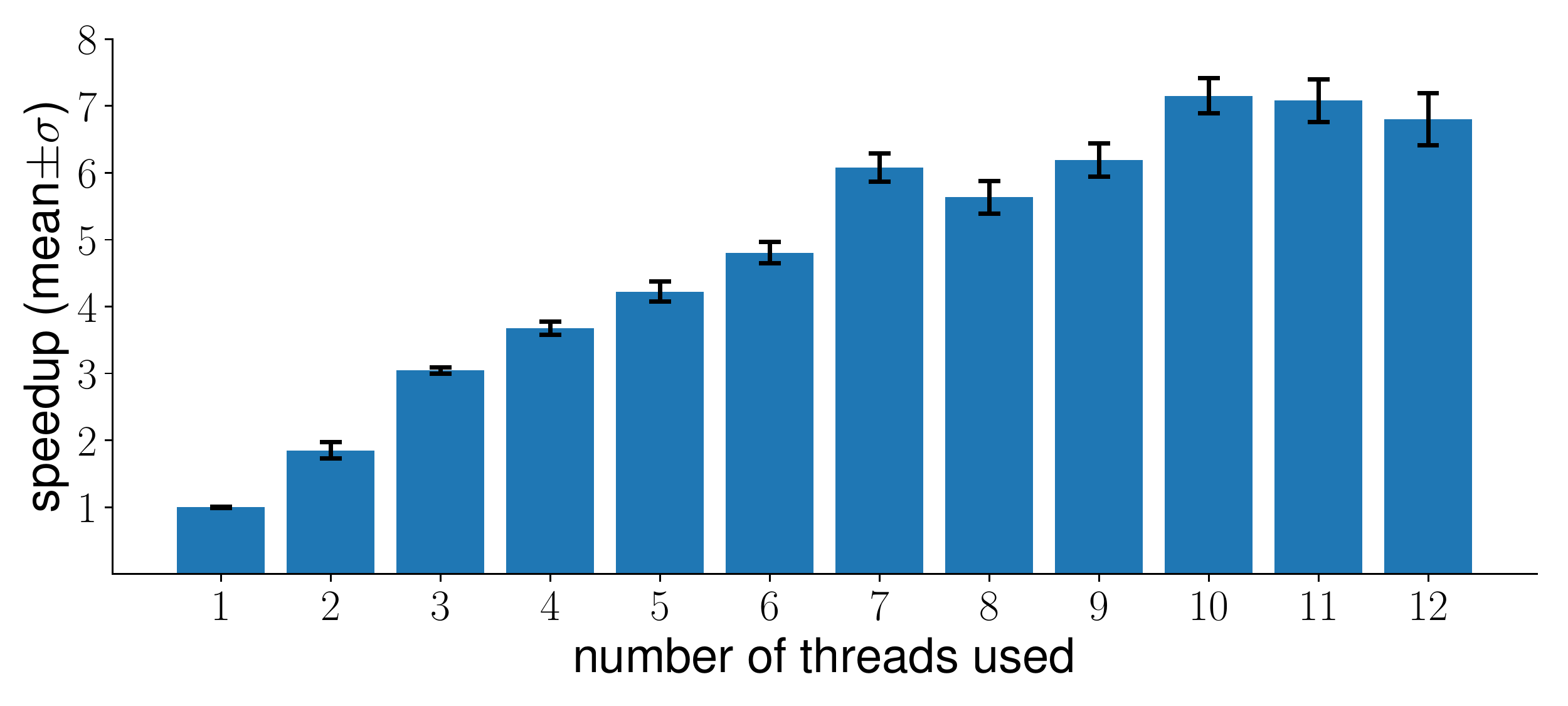}
	\caption{Scaling with number of threads}
	\label{fig:parallelizationplot}
\end{figure}

Due to the above, we show the scaling behavior using a single image. The computations were performed on the workstation for a $512\times 499\times 512$ image with 16-bit precision. We used from 1 to 12 threads taking the mean running time and standard deviation across 20 runs. Figure~\ref{fig:parallelizationplot} shows the speed-up gained using $w$ threads instead of one. In particular using 10 threads is 7.1 times faster than using a single thread.

Table~\ref{tab:huge} shows that the Euler characteristic curve of terabyte scale images can be computed on a single computer with limited memory. Memory usage could be further decreased by changing the chunking scheme. However, the experiments demonstrate that---for the foreseeable future---our implementation is a reasonable trade-off between performance and simplicity.
\begin{table}[!h]
	\centering
	\caption{Running time and memory usage for computations on large 3D images, performed on the 12-core workstation. The voxel values are generated independently from a uniform random distribution in the range $\{0,1,\dots,250\}$. As we already showed, using other images of the same size will exhibit almost identical performance.}
	\label{tab:huge}
	\begin{tabular}{lcrr}
		\hline
		size                 &   threads & \multicolumn{1}{c}{time} & memory   \\
		\hline
		4096$\times$4096$\times$4096& 12 & 1.8 hours & 1.93\,GB\\
		14\,500$\times$14\,500$\times$2\,600  &        24 & 9 hours   & 4.5\,GB    \\
		14\,500$\times$14\,500$\times$2\,600  &        12 & 13 hours  & 2.27\,GB   \\
		10\,000$\times$10\,000$\times$10\,000&         8 & 32 hours  & 3.98\,GB   \\
		\hline
	\end{tabular}
\end{table}
\section{Related Work and Discussion}
We review the work related to computing the Euler characteristic (curve) of images. We embed this in the context of computing other topological descriptors of images, particularly persistence diagrams.

Algorithms related to the Euler characteristic received a lot of attention in image processing~\cite{dyer1980computing,saha1995new,sossa1996computation,ziou2002generating}, starting from the seminal work by Gray~\cite{gray1971local}. Many modern implementations aim at real-time processing of small 2D images~\cite{Snidaro20031533}. Our goal is different, namely handling large multidimensional images. 

Computing other topological descriptors of images is a more recent advancement~\cite{verri1993use,pikaz1997efficient,kaczynski2004computational,wagner2011efficient,robins2011theory,gunther2012efficient,delgado2015skeletonization}, which has however not entered mainstream image processing. Specialized methods for computing persistence diagrams handle 3D images up to $500^3$ voxels~\cite{delgado2015skeletonization,gunther2012efficient} on what we consider commodity hardware. The main limitation is the storage of the entire image in memory. There exist distributed implementations~\cite{bauer2014distributed}, which alleviate the storage problem per machine, but are not specialized to image data, resulting in large overall memory overhead. The largest reported computed instances are in the range of $1000^3$ on 32 server nodes.

Overall it is clear that a specialized, streaming approach is necessary for handling large images. We offer a robust implementation for the Euler characteristic curve, with possible future extensions to other topological descriptors.

We expect that these more complex topological descriptors will be computable for this terabyte scale data in the future but currently we are limited to using Euler characteristic curves. Let us discuss the properties of our algorithmic scheme and mention limitations of our current implementation.

\noindent The advantages of our algorithm are:
\begin{itemize}
	\item It can handle arbitrarily large images on commodity hardware.
	\item It can handle images of arbitrary dimension. 
	\item Linear running time.
	\item Predictable running time and memory usage.
	\item Due to lock-free parallelism, running time scales well with increasing number of threads.
	\item Our algorithm can be easily adapted to a massively-distributed setting using a map-reduce framework~\cite{dean2008mapreduce}.
\end{itemize}

\noindent Some limitations of the current implementation: 
\begin{itemize}
	\item It uses $(3^d-1)$-connectivity. For other types (e.g., 6-connectivity for 2D images), modifications on the algorithm can be made. 
	\item For simplicity we use slices of the image as chunks. For very large images even a one voxel thick slice may not fit into memory. 
	\item For technical reasons we surround each chunk with a second collar of voxels with value $\infty$. Effectively a five voxel thick slice has to fit into memory, which may become a problem for very large images. 
	\item To include the value $\infty$ we may need a larger data type. For example if the input contains all values from 0 to 255 we use a 16-bit data type to store the original values along with an extra value for infinity. 
\end{itemize}
Despite the limitations, our implementation is robust and can handle even the largest data produced by state-of-the-art image acquisition technology. We released this software under the name CHUNKYEuler as open source: \url{https://bitbucket.org/hubwag/chunkyeuler/src}.
\bibliography{Literature}
\bibliographystyle{splncs03}
\end{document}